\documentclass[10pt,twocolumn,letterpaper]{article}

\usepackage[pagenumbers]{cvpr} % To force page numbers, e.g. for an arXiv version

\usepackage[dvipsnames]{xcolor}

\definecolor{cvprblue}{rgb}{0.21,0.49,0.74}
\usepackage[pagebackref,breaklinks,colorlinks,citecolor=cvprblue]{hyperref}
\definecolor{Gray}{gray}{0.9}

\def\paperID{4715} % *** Enter the Paper ID here
\def\confName{CVPR}
\def\confYear{2024}

\title{Retrieval-Augmented Egocentric Video Captioning}

\author{Jilan Xu$^{1,5}$ ~Yifei Huang$^{2,5}$ ~Junlin Hou$^{1}$ ~Guo Chen$^{3,5}$ ~Yuejie Zhang$^{1}$\thanks{Corresponding author} ~Rui Feng$^{1}$ ~Weidi Xie$^{4,5}$ 
\\
$^1$Fudan University
$^2$The University of Tokyo
$^3$Nanjing University \\
$^4$Shanghai Jiao Tong University
$^5$Shanghai AI Laboratory 
\\
}

\begin{document}
\maketitle
\begin{abstract}
Understanding human actions from videos of first-person view poses significant challenges. Most prior approaches explore representation learning on egocentric videos only, while overlooking the potential benefit of exploiting existing large-scale third-person videos. 
In this paper, 
(1) we develop \textbf{EgoInstructor}, a retrieval-augmented multimodal captioning model that automatically retrieves semantically relevant third-person instructional videos to enhance the video captioning of egocentric videos,
(2) for training the cross-view retrieval module, we devise an automatic pipeline to discover ego-exo video pairs from distinct large-scale egocentric and exocentric datasets,
(3) we train the cross-view retrieval module with a novel \textbf{EgoExoNCE} loss that pulls egocentric and exocentric video features closer, 
by aligning them to shared text features that describe similar actions,
(4) through extensive experiments, our cross-view retrieval module demonstrates superior performance across seven benchmarks. Regarding egocentric video captioning, EgoInstructor exhibits significant improvements by leveraging third-person videos as references. 
Project page is available at \url{https://jazzcharles.github.io/Egoinstructor/}
\end{abstract}

%\weidi{menton we will release code and models for future research.}
%\jl{PLACEHOLDER:Semantic segmentation aims at clustering the pixels into different groups and assigning a semantic label to each group. This task benefits a wide range of real-world scenarios, including autonomous driving, computer-aided diagnosis and satellite image analysis. Despite promising results \cite{fcn,unet,maskformer}, the development of existing approaches is mainly impeded for two reasons: (1) Numerous annotation cost. Extensive manual pixel-wise annotations are required for training segmentation models; (2) Closed-set category. The model is restricted to the closed-set of object categories and fails to segment objects of novel classes. Whenever a new dataset comes, the model requires fully-supervised re-training. Both of which hinder the practical development of the state-of-the-art segmentation approaches.(1) Numerous annotation cost. Extensive manual pixel-wise annotations are required for training segmentation models; (2) Closed-set category. The model is restricted to the closed-set of object categories and fails to segment objects of novel classes. Whenever a new dataset comes, the model requires fully-supervised re-training. Both of which hinder the practical development of the state-of-the-art segmentation approaches. }
%%%%%%%%% BODY TEXT

\begin{figure}[t]
\centerline{\includegraphics[width=\columnwidth]{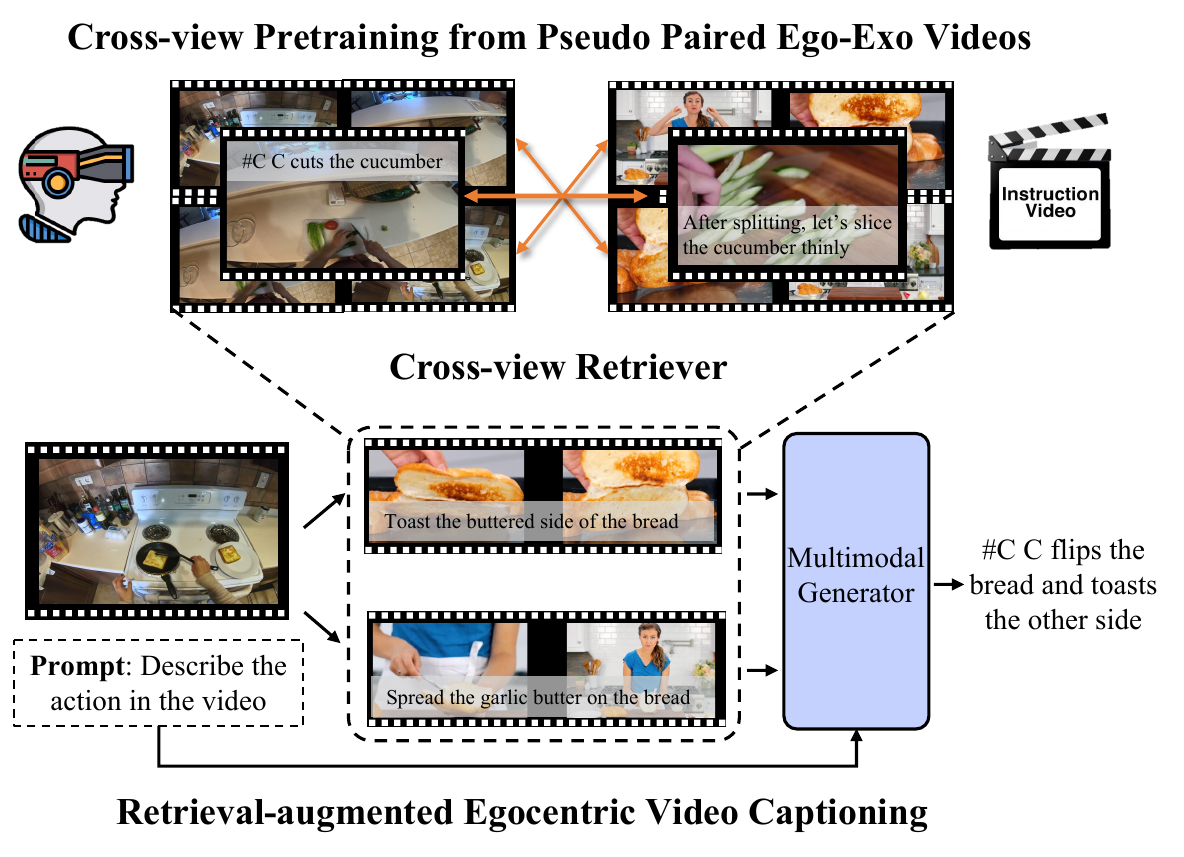}}
\vspace{-5pt}
\caption{EgoInstructor is a retrieval-augmented multimodal captioning model that retrieves relevant exocentric videos as references to generate the caption of egocentric videos. The cross-view retrieval ability is enabled by training on automatically constructed large-scale pseudo paired ego-exo videos.}
\vspace{-10pt}
\label{fig:teaser}
\end{figure}

% \vspace{-2pt}
\section{Introduction}
\label{sec:intro}

Recently, egocentric video understanding is getting increasing attention 
in the vision community, for example, on action recognition~\cite{li2015delving,epic,charadesego,hiervl}, detection~\cite{actionformer,egoonly}, video-text retrieval~\cite{egovlp,lavila}, grounding~\cite{egovlp,naq,internego4d} and gaze estimation~\cite{li2013learning,lai2023eye}.
As egocentric videos are normally recorded from a ``first-person'' view, reflecting activities engaging with the environment, it plays a paramount role in deploying vision models into real-world scenarios, such as robotics~\cite{kumar2020learning} and augmented reality~\cite{nagarajan2020ego}.

As preliminary baseline solutions for egocentric video understanding, 
early works have directly trained models on the egocentric videos~\cite{yamada2012attention,li2013learning,li2015delving,epic},
however, such approaches are limited by dataset scale and have overlooked the benefit of exploiting the large corpus of third-person (exocentric) videos. To leverage the information in exocentric videos, some researchers have tried to directly transfer models learnt from third-person videos~\cite{epicfusion,li2018eye,lsta,egoexo}. 
Due to the discrepancy in terms of recording perspective, camera motion, 
video continuity, and domain shift, the representations learnt from exocentric videos may not be optimal for egocentric videos. 
Another line of research involves the joint learning of egocentric and exocentric video representations~\cite{ardeshir2018exocentric,xu2018joint,ho2018summarizing,charadesego,ae2}. However, training these models would require time-synchronised ego- and exo-video pairs collected in the same environment~\cite{mmac,charadesego,h2o,assembly101}.

In this paper, we take inspiration from the fact that humans are born with the ability of learning by observation, {\em i.e.}, the ability of observing third-view demonstrations and subsequently generalising towards egocentric perspective in different environment~\cite{premack1978does,rizzolatti2004mirror,hodges2007modelled}.
For a more specific example, people often seek answers to ``How To" questions and watch third-person instructional videos before attempting tasks they are unfamiliar with, {\em e.g. how to make a cheesebeef sandwich}. When engaging in the task, people also recall key procedures, visual demonstrations and detailed instructions from instructional videos. We explore retrieval-augmented egocentric video captioning, an alternative way for transferring knowledge from exocentric videos to enhance egocentric video understanding.

Specifically, as shown in Fig.~\ref{fig:teaser}, our goal is to learn a unified ego-exo representation space that allows for explicitly retrieving relevant third-view videos available on the Internet ({\em e.g.} from HowTo100M~\cite{howto100m}), to assist video captioning in the egocentric perspective. This is achieved by first introducing an automated pipeline to generate pseudo ego-exo video pairs at scale by aligning the captions that describe similar actions. Subsequently, the paired videos are employed to train a cross-view retrieval module via a novel EgoExoNCE loss, which aligns both egocentric and exocentric video features with shared text features describing similar action semantics. With this cross-view retrieval capability, we develop a retrieval-augmented egocentric video captioning model by leveraging the retrieved exocentric videos as references for caption generation.

As a consequence, we evaluate the cross-view retrieval module on seven benchmarks, {\em e.g.}, EK100 Multi-instance Retrieval~\cite{epic}, Ego4d Multiple Choice Questions~\cite{ego4d}, YouCook2 video-text retrieval~\cite{youcook} and CharadesEgo cross-view video retrieval~\cite{charadesego}, all of which demonstrate superior performance. Regarding egocentric video captioning on Ego4d~\cite{ego4d}, we show the benefits of leveraging semantically related exocentric instructional videos.

\section{Related Work}
\noindent\textbf{Egocentric video understanding.} 
The unique viewpoint of egocentric videos poses a broad range of challenges in human activity analysis~\cite{survey}, including action recognition and detection~\cite{epic,epicfusion,egoexo,egoonly,actionformer}, human pose estimation~\cite{jiang2017seeing,ng2020you2me}, gaze estimation~\cite{yamada2012attention,li2013learning,huang2018predicting}, and captioning~\cite{nakamura2021sensor,kang2021video}. 
Recently, the introduction of the Ego4d~\cite{ego4d} dataset has prompted a series of studies for representation learning in egocentric videos~\cite{egovlp,lavila,helpinghands,hiervl}. 
For instance, EgoVLP~\cite{egovlp} performed contrastive learning over paired egocentric videos and narrations. 
LaViLa~\cite{lavila} learnt egocentric video-language representations by using pseudo-labelled egocentric videos. Despite the progress in egocentric video understanding, these models lack enough generalisation ability on exocentric videos. In contrast, our cross-view retrieval module is designed to learn a unified ego-exo video-language representation space.

\vspace{3pt}
\noindent\textbf{Ego-exo video understanding.} 
Ego-exo video understanding models involve both egocentric and exocentric videos to solve a variety of vision tasks~\cite{wu2013cross,fan2017identifying,ardeshir2018exocentric,xu2018joint,ego2top,ho2018summarizing,charadesego,egoexo}. 
The majority of these approaches explore shared visual clues from synchronous egocentric and exocentric data recorded simultaneously ~\cite{charadesego,fan2017identifying,assembly101}, 
which poses additional difficulty and cost in data collection. 
Subsequently, versatile methods are developed without relying on paired egocentric and exocentric videos. EgoExo~\cite{egoexo} transferred exocentric pre-trained model to egocentric videos by mining egocentric pseudo labels. AE2~\cite{ae2} temporally aligned the representations of the unpaired egocentric and exocentric videos via Dynamic Time Warping~\cite{dtw}. 
Sum-L~\cite{wang2023learning} created cross-view pseudo pairs from egocentric dataset~\cite{epic,charadesego} and exocentric dataset~\cite{jia2020lemma} using the ground-truth semantic category labels. In contrast, our method learns joint ego-exo representations on unpaired egocentric~\cite{ego4d} and exocentric~\cite{howto100m} datasets by leveraging captions/narrations and moves beyond vision-only~\cite{egoexo,ae2,wang2023learning} tasks to video-language understanding.

\vspace{3pt}
\noindent \textbf{Retrieval-augmented models.}
Retrieval-augmented language models in the NLP community aim at retrieving external knowledge to enhance performance in various NLP tasks~\cite{guu2020retrieval,lewis2020retrieval,borgeaud2022improving,asai2023retrieval,ram2023context}. Motivated by these approaches, recent progresses in vision-language domain also retrieve semantically related samples to improve the performance, including image recognition~\cite{long2022retrieval,iscen2023retrieval}, captioning~\cite{ramos2023smallcap,sarto2022retrieval}, visual question answering~\cite{yang2022empirical,chen2022murag}, image generation~\cite{chen2022re,yasunaga2023retrieval} and vision-language pre-training~\cite{xu2021videoclip}. While most of these efforts primarily focus on image domain and employ a pre-trained CLIP~\cite{clip} model for cross-modal retrieval, such approaches are sub-optimal for video data, 
especially in the case between egocentric and exocentric views. 
In contrast, our approach involves training a cross-view retrieval module tailored for retrieving exocentric videos for egocentric video understanding.

\begin{figure*}[t]
\centerline{\includegraphics[width=\textwidth]{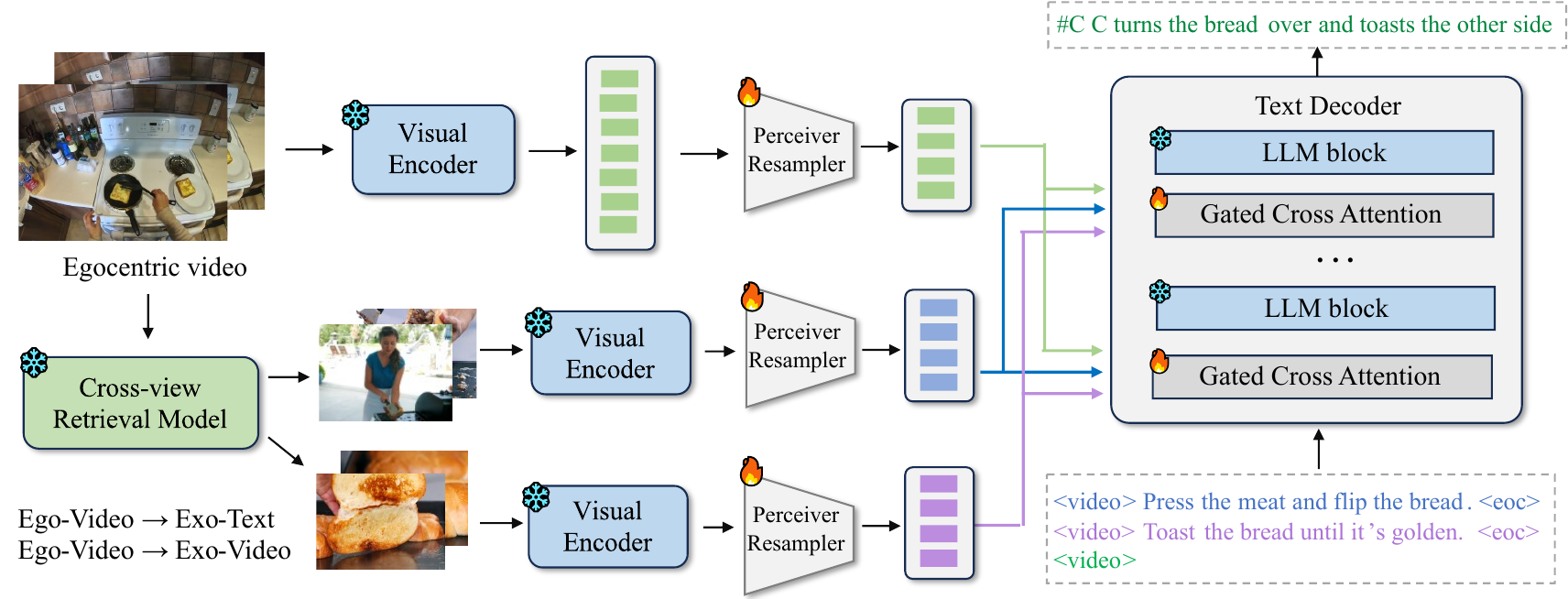}}
\vspace{-5pt}
\caption{An overview of our EgoInstructor. Given an egocentric video, we first retrieve relevant exocentric instructional videos using a frozen cross-view retrieval module pre-trained on pseudo ego-exo pairs generated automatically. The multimodal captioning model (consisting of a visual encoder, a perceiver resampler, and a text decoder.) takes the egocentric video and the retrieved videos and captions as references, and generates the caption of the ego-video.}

\vspace{-10pt}
\label{fig:model}
\end{figure*}

\section{Methodology}

The overall architecture of our proposed EgoInstructor is shown in Fig.~\ref{fig:model}. Given an egocentric video $v^{\text{ego}}$, 
our model first retrieves $K$ semantically relevant third-view instructional videos and associated texts with the cross-view retrieval, 
{\em i.e.}, $\{x^{\text{exo}}_1, \dots, x^{\text{exo}}_K\} = \Phi_{\text{retrieve}}(v^{\text{ego}})$, where $x^{\text{exo}}_i=(v^{\text{exo}}_i,t^{\text{exo}}_i)$ includes both video and associated text narrations. Then, the multimodal captioning model takes both the ego-video and retrieved samples to generate the text description,
$t^{\text{ego}} = \Psi_{\text{generate}}(v^{\text{ego}},  \{x^{\text{exo}}_1, \dots, x^{\text{exo}}_K\})$. In the following sections, we start by describing the procedure for training the cross-view retrieval module with unpaired data (Sec.~\ref{subsec:pair}), then followed by the full retrieval-augmented multimodal captioning model~(Sec.~\ref{subsec:gen}).

\subsection{Cross-view Visual Representation Alignment}
\label{subsec:pair}

Our goal in this step is to train a cross-view retrieval module that associates the ego- and exo-videos sharing the same semantics, using videos with only caption or narrations, {\em i.e.}, without manual annotation for pairing ego-exo videos. Specifically, we adopt \textbf{egocentric} videos from Ego4D~\cite{ego4d} with 4M manually labelled captions, and \textbf{exocentric} videos from HowTo100M~\cite{howto100m}, containing detailed instructions for performing daily procedural activities.

\subsubsection{Architecture Detail}

\noindent\textbf{Egocentric visual-language representation.}
For egocentric video clips, we extract representations via a CLIP-style dual encoder architecture, adopting a frozen VideoMAE~\cite{videomae} as the visual encoder,  $z_i^{\text{ego}}=f^{\text{ego}}(v_i^{\text{ego}})\in\mathbb{R}^{T\times d}$ with $T$ frames;
and a BERT model~\cite{bert} as the text encoder $u_i^{\text{ego}}=f^\text{text}(t_i^{\text{ego}})\in\mathbb{R}^{d}$.
Both visual and text encoders have been pre-trained on Ego4d video-text pairs.

\vspace{3pt}
\noindent\textbf{Exocentric visual-language representation.} 
For exocentric video clip and its instruction from HowTo100M dataset,
the visual representation is computed using a frozen InternVideo~\cite{internvideo} video encoder. This encoder is a Uniformer~\cite{uniformerv2} model pre-trained on large corpus of exocentric video-text pairs~\cite{internvideo}, $z_i^{\text{exo}}=f^{\text{exo}}(v_i^{\text{exo}})\in\mathbb{R}^{T\times d}$. The text features are also computed with the same BERT encoder used for ego-text processing, $u_i^{\text{exo}}=f^\text{text}(t_i^{\text{exo}})\in\mathbb{R}^{d}$.

\begin{figure}[t]
\centerline{\includegraphics[width=\columnwidth]{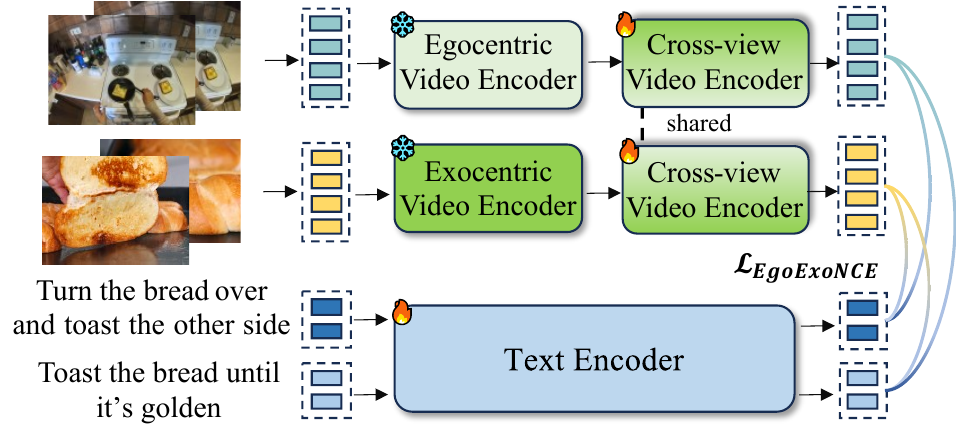}}
\vspace{-5pt}
\caption{Our cross-view retrieval module trained via EgoExoNCE loss. We keep the egocentric and exocentric video encoders frozen and train the cross-view video encoder and text encoder.}
\vspace{-10pt}
\label{fig:retrieval}
\end{figure}

\begin{figure*}[t]
\centerline{\includegraphics[width=\textwidth]{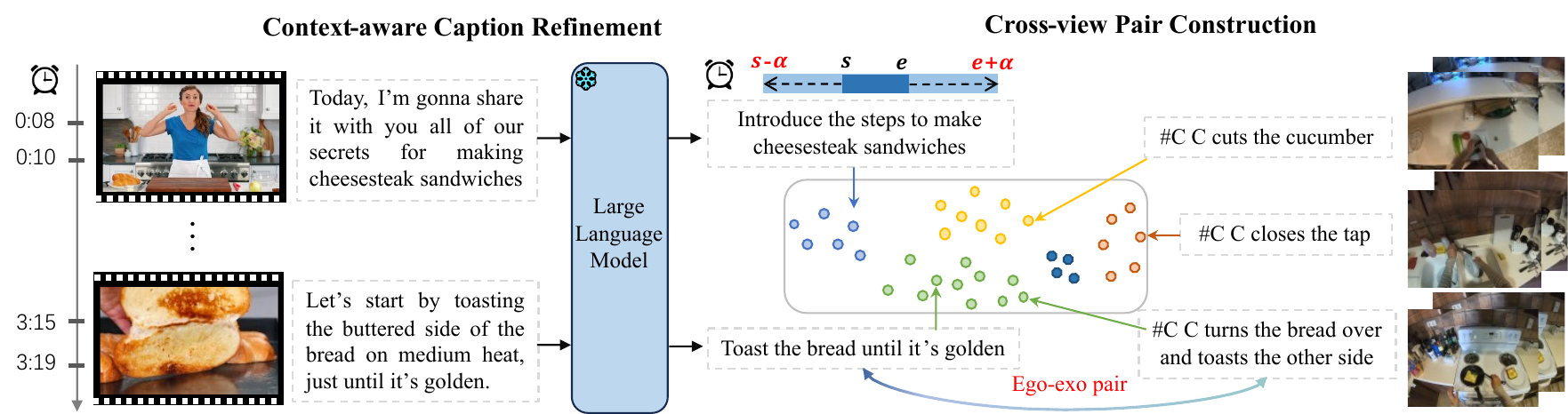}}
\vspace{-5pt}
\caption{An illustration of context-aware caption refinement (left) and cross-view pair construction (right). The ASR transcripts of instructional videos are concatenated and refined by a LLM to match the descriptive style of manually labelled captions in Ego4d. We construct the ego-exo pairs by choosing the ego and exo captions that describe the similar action (e.g., toast the bread).}
\vspace{-10pt}
\label{fig:pair}
\end{figure*}

\vspace{3pt}
\noindent\textbf{Cross-view video encoder.}
With computed visual-language representations, 
we train a cross-view video encoder $\phi(\cdot)$ that takes either egocentric or exocentric video feature as input, and maps them into an embedding space, such that the videos with similar semantics are close together regardless of the shooting perspective. 
We obtain $\hat{z}_i^{\text{ego}}=\phi(z_i^{\text{ego}})\in\mathbb{R}^{d}$ and $\hat{z}_i^{\text{exo}}=\phi(z_i^{\text{exo}})\in\mathbb{R}^{d}$ with the temporal dimension being aggregated with average pooling. The cross-view retrieval module is shown in Fig.~\ref{fig:retrieval}.

\subsubsection{Automatic Ego-Exo Pair Generation}
For training the described cross-view retrieval module, 
we propose a scalable approach that exploits natural languages as a bridge to pair videos from different viewpoints, {\em i.e.}, grouping egocentric and exocentric videos if their corresponding language narrations and instructions are highly relevant. In practice, unlike egocentric videos from Ego4D that have manually annotated captions, instructions for HowTo100M videos are normally acquired from ASR transcripts~\cite{whisperx}. 
This incurs two issues: 
(i) the transcript may not be of descriptive style, containing redundancy or ambiguity, as depicted in Fig.~\ref{fig:pair}~(left); 
(ii) the transcript may not be well aligned with the visual signals. 
There might be greetings from the speaker or inaccuracies in timing,
{\em e.g.}, describing the action after performing it, as discussed in~\cite{han2022temporal}.

\vspace{4pt} 
\noindent \textbf{Caption refinement via large language model.}
We transform the ASR transcript into similar style as those of Ego4D captions with a large language model~(LLM). 
Specifically, given an instructional video from HowTo100M,
{\em i.e.}, $x^{\text{exo}}=\{(v_j,t_j,s_j,e_j)\}_{j=1}^ M$, 
where each narration $t_j$ is annotated with a start time $s_j$ and end time $e_j$, we refine the ASR transcript by prompting the LLM to only capture the main actions in the transcript and output descriptive captions:
\begin{equation*}
\small
    [t_1',..., t_M'] = \text{LLM}([\textbf{Prompt},(t_1,s_1,e_1),...,(t_M,s_M,e_M)]),
\end{equation*}
where \textbf{Prompt} is composed of task instructions, 
a list of 10 caption rules followed by example cases. 
Please refer to \textbf{Supplementary} for the exact prompt. 
Compared with clip-wise processing of each transcript individually, video-level refinement enables the LLM to infer the long-term goal of the activity within the context, and manages to replace the pronouns with specific objects deduced from other transcripts for better visual-text alignment. Additionally, we extend the temporal boundary of the narration to $[s-\alpha, e+\alpha]$ by $\alpha$ seconds, to mitigate the temporal shift problem. 
As shown in Fig.~\ref{fig:pair}~(middle), the ASR transcripts are effectively transformed into shorter yet more informative sentences. 

\vspace{4pt}
\noindent\textbf{Pairing ego-exo videos via language alignment.} 
In addition to refining the captions of HowTo100M videos, 
we also remove the character indicator in Ego4D narrations, 
{\em e.g.}, the \textit{`\#C'} in captions.
Till this point, the egocentric and exocentric narrations share similar textual formats, we then establish pairings via a two-step procedure. 
{\em First}, we conduct an initial categorisation based on the daily activity scenarios, such as food and entertaining, crafts, gardening, 
{\em etc}. Egocentric and exocentric videos belonging to the same scenario are grouped together; 
{\em Second}, we extract nouns and verbs using Spacy~\cite{spacy} from each sentence as they collaboratively determine the semantics of the activity in the video. For each egocentric video, the exocentric video(s) with the highest overlaps in terms of nouns and verbs are selected as the paired sample(s). Note that, we have also experimented by computing sentence similarities directly, however, it is sub-optimal than explicitly comparing entities.

\vspace{4pt}
\noindent\textbf{Long-form video-text pairs construction.} 
Till here, we have obtained paired ego-exo short video clips, 
where the duration of video clips are determined by the timestamp of the captions. 
As both Ego4d and HowTo100M consist of long videos lasting minutes, 
training on short video-text pairs may not fully exploit the information in long videos. 
To address this, we propose a simple strategy to better utilise long-form information. 
In specific, as the narrations/captions are dense, we create `fake' long-form video-text pairs by cropping longer video clips and summarising their accompanied narrations into one sentence via the LLM.
These pseudo long-form pairs are integrated into the original data for training, and pairing long-form ego-exo video pairs can be executed similarly as before. 

\subsubsection{Retrieval Module Training and Inference}
For training cross-view retrieval, we introduce a novel contrastive loss, termed as EgoExoNCE loss. Specifically, the proposed loss incorporates three variants of positive pairs: (1) the original video-text pairs $(\hat{z}_i^{\text{ego}},u_i^{\text{ego}})$ and $(\hat{z}_i^{\text{exo}},u_i^{\text{exo}})$ as adopted in the InfoNCE loss~\cite{clip}; (2) cross-view pairs mined via language alignment, 
{\em i.e.}, $(\hat{z}_i^{\text{ego}},u_i^{\text{exo}})$ and $(\hat{z}_i^{\text{exo}},u_i^{\text{ego}})$, as both $u_i^{\text{ego}}$ and $u_i^{\text{exo}}$ represent similar actions in the video; 
(3) all pairs that share at least one noun or verb \textit{regardless of views}. The remaining pairs in each mini-batch are considered as negative pairs.

At training time, assume we have $\mathcal{B}$ samples in each mini-batch, 
the positive pair set for the $i$-{th} sample is thus defined as:

\begin{equation*}
\small
\begin{aligned}
    &\mathcal{P}_i=\{(\hat{z}_i^{\text{ego}},u_i^{\text{ego}}),(\hat{z}_i^{\text{exo}},u_i^{\text{exo}})\}\cup \{(\hat{z}_i^{\text{ego}},u_i^{\text{exo}}),(\hat{z}_i^{\text{exo}},u_i^{\text{ego}})\}\cup \\
    &\{(\hat{z}_i^{\text{ego}},u_j^c),(\hat{z}_i^{\text{exo}},u_j^c)|\mathcal{N}(t_i^c)\cap\mathcal{N}(t_j^c)\neq\emptyset,\mathcal{V}(t_i^c)\cap\mathcal{V}(t_j^c)\neq\emptyset \},
\end{aligned}
\end{equation*}
where $j\in\mathcal{B}, c\in\{\text{ego},\text{exo}\}$, $\mathcal{N}(t)$ and $\mathcal{V}(t)$ represent the sets of nouns and verbs in text $t$. 
The video-to-text EgoExoNCE loss $\mathcal{L}^{\text{v2t}}_{\text{EgoExoNCE}}$ can be calculated as:
\begin{equation}
\small
\mathcal{L}^{\text{v2t}}_{\text{EgoExoNCE}}=-\frac{1}{|\mathcal{B}|} \sum_{i\in \mathcal{B}}\log \frac{\sum_{(\hat{z}_i,u)\in\mathcal{P}_i}\left\langle{\hat{z}_i},u\right\rangle}{\sum_{j\in\mathcal{B}}\left\langle{\hat{z}_i^{\text{ego}}},u_j\right\rangle+\left\langle{\hat{z}_i^{\text{exo}}},u_j\right\rangle},
\end{equation}
where $\left\langle z,u\right\rangle=\exp(z\textsuperscript{T}u/\tau)$ and $\tau$ is the temperature. The total loss $\mathcal{L}_{\text{EgoExoNCE}}$ is the sum of video-to-text loss and symmetrical text-to-video loss.
Note that, despite the training procedure only encourages cross-view, 
cross-modal retrieval, in fact, the ego-exo video embeddings are also implicitly pulled closer, by aligning to the shared semantics of the actor's action. 

At inference time, given an egocentric video and a candidate set of exocentric videos with associated texts, 
we first calculate egocentric video representation~($\hat{z}^{\text{ego}}$) and exocentric video and text representations~$(\hat{z}^{\text{exo}},u^{\text{exo}})$.
Then, for the egocentric video, we choose the exocentric sample with the highest averaged similarity, considering both (1) ego-video to exo-video similarity and (2) ego-video to exo-text similarity, which is defined as:
\begin{equation}
    \mathop{\arg\max}_{(v^{\text{exo}},t^{\text{exo}})\in\mathcal{C}}\frac{1}{2}(\langle\hat{z}^{\text{ego}},\hat{z}^{\text{exo}}\rangle+\langle\hat{z}^{\text{ego}},u^{\text{exo}}\rangle).
\end{equation}

\subsection{Retrieval-augmented Captioning}
\label{subsec:gen}

In this section, we detail the proposed retrieval-augmented model for egocentric video captioning. In addition to the target video as input, retrieved instructional videos and associated texts are also incorporated as conditions for generating captions. Next, we present three key modules in our captioning model, as shown in Fig.~\ref{fig:model}.

\vspace{3pt}
\noindent\textbf{Visual encoder.} 
We adopt a frozen CLIP vision transformer encoder that takes the concatenation of exo- and ego-videos $v=[v^{\text{exo}}_1;...;v^{\text{exo}}_K;v^{\text{ego}}]\in\mathbb{R}^{(K+1)\times T\times C\times H\times W}$ as input, tokenises each frame into $N$ image patches, and compute dense visual representation per frame, $z=\Phi_\text{v}(v)\in\mathbb{R}^{(K+1)\times T\times N\times d}$. Here, [$\cdot$] refers to concatenation.

\vspace{2pt}
\noindent\textbf{Perceiver resampler} serves as a bottleneck module, that models the correspondence among the visual features and squeezes the visual features to a bottleneck feature with fixed length $L$, typically $L\ll T\times N$. This is achieved with Transformer Decoder using fixed-length of learnable query tokens $q\in\mathbb{R}^{L\times d}$,
that cross-attend to the visual features, 
{\em i.e.}, $\hat{q}=\Phi_\text{p}(q, [z+\textbf{pe};q])\in\mathbb{R}^{(K+1)\times L\times d}$. Here, $\textbf{pe}$ denotes temporal positional embedding.

\vspace{3pt}
\noindent\textbf{Text decoder} is a visually conditioned LLM. 
It takes captions of the retrieved samples as input prompt and cross-attends to visual features for caption generation. 

The text input prompt $\textbf{t}_{\text{p}}$ is formed by stacking captions for all exocentric videos, which is denoted as:

\begin{equation*}
    \textbf{t}_{\text{p}}=[\langle \text{video}\rangle t_1^{\text{exo}}\langle \text{eoc}\rangle;...; 
    \langle \text{video}\rangle t_K^{\text{exo}}\langle \text{eoc}\rangle;\langle\text{video}\rangle],
\end{equation*}
where $\langle \text{video}\rangle$ and $\langle \text{eoc}\rangle$ are special media tokens representing the video and end-of-chunk in text format, respectively.

In particular, several gated cross-attention modules~\cite{flamingo,openflamingo} are inserted between the original LLM blocks. Compared with the standard LLM block, the gated cross-attention module exhibits two key variations: (1) the cross-attention layer takes text features as query and visual features as keys and values, enabling visually conditioned generation. Note that, each text token only cross-attends to the visual feature of its corresponding video. (2) A $\tanh$ gating mechanism is added. The output of cross-attention layer and feed-forward-network in each gated cross-attention block is multiplied by a $\tanh$ layer with zero-initialisation. This ensures that the decoder is initially equivalent to the original LLM, thereby improving the training stability. The captioning model is trained by optimising the standard cross-entropy loss for each predicted token.

\section{Experiments}
\subsection{Experimental Setups}
\label{sec: impl}
\begin{table}[t]
\centering
\resizebox{\columnwidth}{!}{
  \begin{tabular}[t]{lcccc}
    \toprule
    Dataset & Evaluation Metrics & \#Len(s) & \#Num \\
    \midrule
    \emph{Egocentric benchmark}  \\
    EK100 MIR~\cite{epic} & mAP, nDCG & 3.7 & 9668\\
    EgoMCQ~\cite{ego4d} & inter-/intra-video acc. & 34.2 & 39751 \\
    SummMCQ~\cite{ego4d} & inter-video acc. & 181.6 & 1614 \\
    \midrule
    \emph{Exocentric benchmark}  \\
    YouCook2-Clip~\cite{youcook}  & R@1, R@5, R@10 & 19.7 & 3350 \\
    YouCook2-Video~\cite{youcook} & R@1, R@5, R@10 & 212.4 & 436 \\
    \midrule
    \emph{Ego-Exo benchmark}  \\
    CharadesEgo~\cite{charadesego}  & Ego2Exo/Exo2Ego Avg. R@1/@5/@10 & 22.7 & 145 \\
    EgoLearner-MCQ & Ego2Exo/Exo2Ego acc. & 5.1 & 1951 \\
    \midrule
    \midrule 
    \emph{Video Captioning} \\
    Ego4d cooking~\cite{ego4d} & BELU-4, METEOR, ROUGE-L, CIDER & 0.7 & 7161 \\
    EgoLearner & BELU-4, METEOR, ROUGE-L, CIDER & 7.4 & 1089 \\
    \bottomrule
  \end{tabular}
}
\vspace{-5pt}
\caption{Datasets for evaluation. We list evaluation metrics, average clip/video length (\#Len), and the number of test samples.} %Datasets with average clip length less than a minute are categorized as short-term benchmarks.}
%\vspace{-0.4cm}
\vspace{-10pt}
\label{tab:dataset}
\end{table}

\begin{table*}[t]
\centering
\resizebox{\textwidth}{!}{
  \begin{tabular}[t]{lcccccccccccccc}
    \toprule
    &
    &
    &
    & 
    \multicolumn{4}{c}{Ego-centric Benchmark} &
    \multicolumn{3}{c}{Exo-centric Benchmark} &
    \multicolumn{4}{c}{Ego-Exo Benchmark} \\
    
    \cmidrule(r){5-8} \cmidrule(r){9-11} \cmidrule(r){12-15} 
    ID&Loss&Caption&View& \multicolumn{2}{c}{EK100-MIR} & 
    \multicolumn{2}{c}{EgoMCQ} & \multicolumn{3}{c}{YouCook2 Clip$\rightarrow$ Text}  & \multicolumn{2}{c}{EgoLearner}&\multicolumn{2}{c}{CharadesEgo}\\
    \cmidrule(r){5-6} \cmidrule(r){7-8} \cmidrule(r){9-11} \cmidrule(r){12-13}\cmidrule(r){14-15}

    &&&& mAP & nDCG & Inter Acc. & Intra Acc. & R\texttt{@}1 & R\texttt{@}5 &  R\texttt{@}10 & Ego2Exo & Exo2Ego & Ego2Exo & Exo2Ego\\
    
    \cmidrule(r){1-4} \cmidrule(r){5-6} \cmidrule(r){7-8} \cmidrule(r){9-11} \cmidrule(r){12-13} \cmidrule(r){14-15} 

    1&InfoNCE & - & Ego  & 29.9 & 33.9 &  91.5 & 54.6 & 0.1 & 0.5 & 0.6 & 29.4& 22.6 & 6.1 & 7.4\\
    2&InfoNCE & ASR & Exo & 6.7 & 12.3 &  35.0 & 22.6 & 21.7 & 48.1 & 61.2 & 24.6& 21.9 & 1.8 & 4.1\\
    \midrule
    \midrule
    3&InfoNCE & ASR & Ego+Exo & 28.1 & \textbf{32.1} & 91.5 &51.8&21.3 & 45.4 & 57.3  & 38.7 & 29.9 & \textbf{18.3} & 7.6\\ 
    4&InfoNCE & Refined Cap. & Ego+Exo & \textbf{28.4} & 32.0& \textbf{92.3} & \textbf{53.2} & \textbf{23.3}&\textbf{49.2}&\textbf{60.7}
     & \textbf{46.8}&\textbf{30.1} &17.8&\textbf{9.8}\\
    \midrule
    \midrule
    5&EgoNCE & ASR & Ego+Exo & 29.0 & 33.1 & 91.5 & 52.5 & 21.4 & 45.0 & 57.6 & 39.2 & 30.9& 19.0 & 8.7\\ 
    6&EgoExoNCE & ASR & Ego+Exo & 28.6 & \textbf{33.2} & 91.9 & 51.4 & 21.1 & 45.6 & 57.7  & 41.3 &\textbf{31.3} & 20.8 & 13.6\\
    7&EgoExoNCE & Refined Cap. & Ego+Exo & \textbf{29.2} & 33.1 & \textbf{92.8} &\textbf{ 54.0 }& \textbf{23.7} & \textbf{49.4} &\textbf{ 61.8} &  \textbf{46.4} & 30.5 & \textbf{22.6} & \textbf{15.2}\\
    \bottomrule
  \end{tabular}
}
\vspace{-5pt}
\caption{Ablation study on the exocentric instructional video textual descriptions (original ASR transcripts vs. Refined Caption) and the cross-view alignment loss (InfoNCE~\cite{clip} vs. EgoNCE~\cite{egovlp} vs. EgoExoNCE). We report metrics on short-term benchmarks. The baseline model trained on ego-only/exo-only data (ID-1, ID-2) fails on exo/ego benchmarks due to the discrepancy between views.}
%\vspace{-0.4cm}
\vspace{-10pt}
\label{tab:egoexonce}
\end{table*}

\noindent\textbf{Training datasets.} 
For egocentric videos, we use the public Ego4d dataset~\cite{ego4d}, consisting of 3670 hours of daily activity videos. 
Following prior works~\cite{egovlp,lavila}, 
videos belonging to the validation and test sets are excluded during training, resulting in 4.3M video-text pairs for training~\cite{lavila}. 
For exocentric videos, we adopt the large-scale instructional video dataset HowTo100M~\cite{howto100m}, containing more than 1M videos from YouTube. 
We use the annotation from TAN~\cite{han2022temporal} including 30M video and ASR transcript pairs with better temporally aligned narration timestamps. 
For training the video captioning model, we use the Ego4d~\cite{ego4d} cooking subset, containing 0.55M video clips with manually labelled captions. All video clips are obtained based on the timestamps of the captions/narrations.

\vspace{4pt}
\noindent\textbf{Downstream benchmarks.}
We evaluate the zero-shot transfer ability of our \textbf{retrieval module} on the validation/test set  using benchmark datasets outlined in Table~\ref{tab:dataset}. Our evaluation includes 7 retrieval benchmarks that cover both egocentric~\cite{epic,ego4d,charadesego} and exocentric videos~\cite{youcook,charadesego}, ranging from short-term to long-term. 
Notably, CharadesEgo and an in-house dataset, termed as EgoLearner, with ground-truth ego-exo pairs are adopted to evaluate model's video-to-video retrieval ability. The retrieval evaluations include video-text, video-video retrieval, and multiple choice questions (MCQ). 
Specifically, MCQ involves retrieving the matched video from 5 candidates from a query. The query is the associated text in EgoMCQ and SummMCQ. 
While in EgoLearner, the video from the other view serves as the query. We report the R@1, R@5 and R@10 for retrieval tasks and Top-1 accuracy for MCQ tasks. For \textbf{video captioning}, we sample 7161 cooking video clips from the Ego4d~\cite{ego4d} validation set. We also choose videos from EgoLearner which contains 1089 video clips with fine-grained video captions. Please refer to the \textbf{Supplementary} for details on each dataset.

\vspace{4pt}
\noindent\textbf{Training and inference pipeline.} 
To begin with, we take egocentric videos from Ego4d and exocentric videos from HowTo100M to construct pseudo ego-exo pairs offline, which are subsequently used to train the cross-view retrieval module. 
Next, we take the videos from the cooking subset of Ego4d, along with the paired exocentric videos and captions, to train the retrieval-augmented captioning model. 
At inference stage, given an egocentric video, we retrieve relevant exocentric videos via the cross-view retrieval module, and use them as references to generate the ego-caption. 
%Regarding the retrieval strategy, we average the similarity of (1) ego-video to exo-text and (2) ego-video to exo-video, and rank the exocentric videos based on the averaged similarity. 

\vspace{4pt}
\noindent\textbf{Implementation details.} 
In the retrieval module, the egocentric video encoder is a VideoMAE-L~\cite{videomae} model pre-trained on Ego4d. The exocentric video encoder is a UniformerV2~\cite{uniformerv2} model with InternVideo~\cite{internvideo} pre-trained weights on third-person video datasets. 
Both video encoders output a 768-d feature vector for an 8-frame input. 
Following~\cite{howto100m,han2022temporal}, we get one feature per second by decoding video at 8 fps. 
To match the average duration of Ego4d and HowTo100M, 
we randomly sample 4 frame features as input to the cross-view video encoder by default. 
In terms of pseudo long-form video-text pairs, we crop long video clips that last between 60 and 300 seconds, covering 20 narrations. 
We employ Stable-Vicuna-v1.5~\cite{vicuna}, an advanced instruction-tuned LLM for both long-form text summarisation and caption refinement.
The cross-view video encoder consists of four Transformer encoder layers~\cite{vaswani2017attention}. The text encoder is a BERT model~\cite{bert}. 
The retrieval module is trained for 5 epochs using the AdamW optimizer with a learning rate of $3\times 10^{-5}$, and the batch size is set to 4096. 

For retrieval-augmented captioning model, the visual encoder is CLIP-L/14 and the perceiver resampler consists of 6-layer transformer blocks with 64 learnable query tokens. 
We adopt two variants of language models, LLaMA-7B~\cite{llama} and MPT-1B~\cite{mpt}. 
For fast convergence, we initialise the captioning model with Otter~\cite{otter} pre-trained weights. We train the model for 3 epochs with an initial learning rate of $1\times 10^{-5}$ and cosine learning rate decay. We set batch size to 8/32 for LLaMA-/MPT-based models. 

\subsection{Experimental Results}
\begin{table*}[t]
\centering
\resizebox{\textwidth}{!}{
  \begin{tabular}[t]{lccccccccccccccc}
    \toprule
    & 
    & 
    & 
    \multicolumn{5}{c}{Ego-centric Benchmark} &
    \multicolumn{4}{c}{Exo-centric Benchmark} &
    \multicolumn{4}{c}{Ego-Exo Benchmark} \\
    \cmidrule(r){4-8} \cmidrule(r){9-12} \cmidrule(r){13-16}
    Method&View&Backbone&\multicolumn{2}{c}{EK100-MIR} & 
    \multicolumn{2}{c}{EgoMCQ} & SummMCQ & 
    \multicolumn{2}{c}{YC2-Clip} & \multicolumn{2}{c}{YC2-Video}
    & \multicolumn{2}{c}{EgoLearner} & \multicolumn{2}{c}{CharadesEgo}
     \\
    &&& mAP & nDCG & Inter Acc. & Intra Acc. & Acc & R\texttt{@}1 & R\texttt{@}5 & R\texttt{@}1 & R\texttt{@}5 & Ego2Exo & Exo2Ego & Ego2Exo & Exo2Ego  \\
    
    \cmidrule(r){1-3} \cmidrule(r){4-5} \cmidrule(r){6-7} \cmidrule(r){8-8} \cmidrule(r){9-10} \cmidrule(r){11-12} \cmidrule(r){13-14} \cmidrule(r){15-16}

    %\rowcolor{Gray}
    HierVL~\cite{hiervl} & Ego & FIT & 18.9 & 24.7 & 90.5 & 52.4 & 95.4$^{\dag}$ & - & - & - & - & - & - & - & -\\
    \midrule
    CLIP~\cite{clip} & Ego+Exo & ViT-L & 18.4&24.0 & \textbf{88.7}&34.6 &92.0 & 20.2&44.0 & 36.0&76.4 & 48.9&41.5 & 50.1 & 49.0 \\
    CLIP + Ours & Ego+Exo & ViT-L & \textbf{19.4}&\textbf{24.5} & 88.3&\textbf{35.0} & \textbf{93.5} & \textbf{23.6}&\textbf{52.8}& \textbf{58.0}&\textbf{90.4}& \textbf{49.0}&\textbf{42.3}&\textbf{59.5}&\textbf{59.1} \\
    \midrule
    
    EgoVLP~\cite{egovlp} & Ego+Exo & TSF-B & 24.7 & 28.4 &  91.2&45.7 & 88.8 & 20.2&44.8 & 39.6&78.0 & 52.0&43.8&\textbf{21.7}&22.9\\
    EgoVLP + Ours & Ego+Exo & TSF-B & \textbf{25.7}&\textbf{29.1} & \textbf{91.9}&\textbf{47.2} & \textbf{91.6} & \textbf{22.7}&\textbf{50.5} & \textbf{59.4}&\textbf{91.3}& \textbf{53.2}&\textbf{46.8}&21.2&\textbf{29.2} \\
    \midrule
    LaViLa~\cite{lavila} & Ego+Exo & TSF-L & 24.3&28.2 & 90.8&45.3 & 89.9 & 19.7&44.5 & 37.6&76.6 & 52.6&\textbf{43.3}&\textbf{37.1}&25.9\\
    LaViLa + Ours & Ego+Exo & TSF-L & \textbf{25.7}&\textbf{29.0} & \textbf{91.2}&\textbf{46.5} & \textbf{94.2}&\textbf{23.2} &\textbf{51.2} & \textbf{61.0}&\textbf{89.7}& \textbf{53.7}&41.8&35.5&\textbf{33.6}\\
    \midrule
    InternVideo~\cite{internvideo} & Ego+Exo & UF-L   &  21.4&25.1 & 88.9&38.7 & 88.6 & 20.4&45.4 & 34.8&73.8 & 50.1&42.2&63.9&62.0\\
    InternVideo + Ours & Ego+Exo & UF-L   & \textbf{25.7}&\textbf{27.6} & \textbf{90.3}&\textbf{39.6} & \textbf{93.4} & \textbf{23.6}&\textbf{52.1} & \textbf{59.6}&\textbf{87.6} & \textbf{52.8}&\textbf{48.3} &\textbf{79.8}&\textbf{70.3}\\
    \midrule
    VideoMAE~\cite{videomae} & Ego+Exo & ViT-L   &  28.1 & 32.1 & 91.5 & 51.8 & 87.1 & 21.3 & 45.4 & 37.6 & 75.4 & 38.7 & 29.9 &18.3 & 7.6\\
    VideoMAE + Ours & Ego+Exo & ViT-L  & \textbf{31.6} &\textbf{ 34.9 }& \textbf{92.7} & \textbf{54.2} & \textbf{90.3} & \textbf{25.5} &\textbf{51.8} & \textbf{73.4} & \textbf{96.8}& \textbf{44.0}& \textbf{30.4} & \textbf{25.1} & \textbf{19.3}\\

    \midrule
  \end{tabular}
}
\vspace{-5pt}
\caption{Effect of our cross-view visual representation alignment over various egocentric video encoders, while fixing the exocentric video encoder as InternVideo~\cite{internvideo}. We report clip/video to text retrieval results on YouCook-Clip/YouCook-Video. $\dag$ indicates different test data (still unavailable), which is not fairly comparable. FIT, TSF and UF refer to Frozen in Time~\cite{frozenintime}, TimeSformer~\cite{timesformer} and UniformerV2~\cite{uniformerv2}.}
%\vspace{-0.4cm}
\vspace{-10pt}
\label{tab:backbone}
\end{table*}

\label{sec:expresult}

In this section, we study the effect of each component for cross-view retrieval and retrieval-augmented captioning.

\subsubsection{Results on Cross-view Retrieval}
We evaluate the effect of caption refinement, EgoExoNCE loss, and different egocentric video encoders in this section. Additional ablations can be found in the Supplementary.

\label{subsubsec:ret_result}

\vspace{4pt}
\noindent\textbf{Effect of caption refinement.}
We show comparison between our refined exocentric video caption with the original ASR transcript across multiple losses in Table~\ref{tab:egoexonce}~(ID-3 vs.~ID-4, ID-6 vs.~ID-7). 
The refined caption leads to consistent improvements on YouCook2 exocentric retrieval, indicating better visual-text alignment. 
Surprisingly, we observe 1\%$\sim$2\% improvement on egocentric benchmark, EgoMCQ. 
We conjecture that the performance gain is from the alignment between egocentric and exocentric caption formats achieved through caption refinement. This alignment allows egocentric videos to align with broader semantically rich captions, thereby enhancing the representation of egocentric video-text relations. 
On cross-view benchmark, {\em i.e.}, EgoLearner video retrieval, 
the accuracy consistently improves in the Ego2Exo setting regardless of the used losses, as semantically similar ego- and exo-videos are better aligned via similar captions. 
In the Exo2Ego setting, however, the model exhibits minimal improvement, probably due to the fact that the egocentric videos in EgoLearner are primarily collected in a similar environment, making it challenging to distinguish these ego-videos.

\vspace{4pt} 
\noindent\textbf{Effect of EgoExoNCE loss.}
To investigate the impact of different loss functions on aligning egocentric and exocentric videos, we compared our EgoExoNCE with two commonly used loss functions, namely, InfoNCE~\cite{clip,lavila} and EgoNCE~\cite{egovlp}. 
InfoNCE aligns each video with its paired caption without considering cross-view information. 
EgoNCE extends this by incorporating videos performing similar actions as positive samples, while different actions in the same video as negative samples. 

As shown in Table~\ref{tab:egoexonce}, 
EgoNCE improves InfoNCE on egocentric benchmarks~(ID-3 vs.~ID-5), {\em i.e.}, EK100-MIR and EgoMCQ, however, 
only marginal improvement is observed on cross-view benchmarks. 
In contrast, EgoExoNCE boosts the accuracy of InfoNCE by an average of 2.0\% and 4.3\% on cross-view benchmarks~(ID-3 vs.~ID-6), 
{\em i.e.}, EgoLearner and CharadesEgo, respectively. 
The results can be further improved by using refined captions~(ID-4 vs.~ID-7). This indicates the superiority of our EgoExoNCE loss in leveraging captions to mine cross-view samples, thereby enhancing the model's video alignment ability. 
In summary, our cross-view visual alignment outperforms the baseline on all benchmarks~(ID-3 vs.~ID-7).  

\begin{figure*}[t]
\centerline{\includegraphics[width=\textwidth]{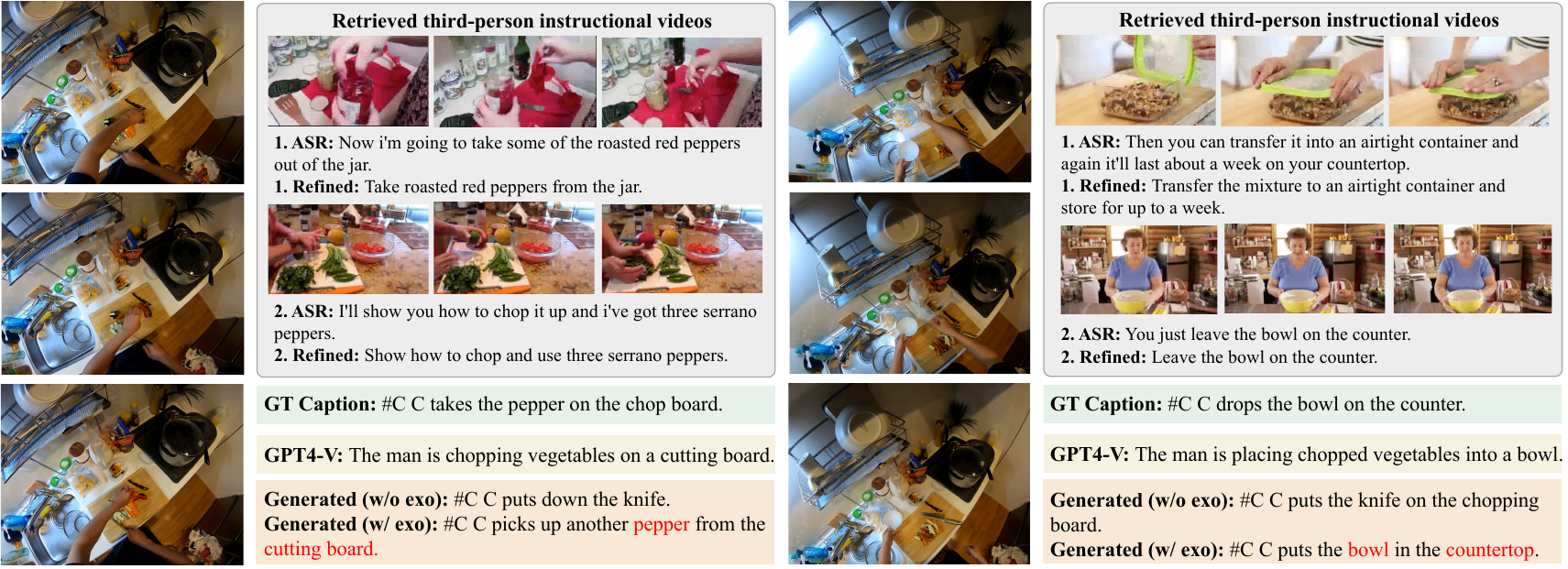}}
\vspace{-5pt}
\caption{\textbf{Visualisation results.} For each egocentric video, we show two retrieved third-person instructional videos and their original ASR/refined captions. By leveraging retrieved exocentric samples, the generated captions capture correct actions and interacting objects.}
\vspace{-5pt}
\label{fig:vis}
\end{figure*}

\vspace{4pt} 
\noindent\textbf{Analysis on egocentric video encoder.} 
We extract egocentric video features using various pre-trained encoders, 
including CLIP~\cite{clip}, EgoVLP~\cite{egovlp}, LaViLa~\cite{lavila}, InternVideo~\cite{internvideo} and VideoMAE~\cite{videomae}, 
while ensuring the consistency of exocentric video features extracted via InternVideo~\cite{internvideo}. The results are shown in Table~\ref{tab:backbone}. 

All the baseline models are trained on both original egocentric and exocentric data via InfoNCE. By adding our cross-view visual representation alignment strategies, 
{\em i.e.,} training on refined paired data with EgoExoNCE loss,

all models exhibit consistent improvements on most performance metrics.
Specifically, on cross-view evaluation benchmarks, decent improvement can be seen on CLIP and InternVideo. For instance, our model improves the baseline InternVideo model by 15.9\% and 7.7\% on CharadesEgo. 

% \vspace{-10pt}
\subsubsection{Results on Retrieval-augmented Captioning}
\label{subsubsec:gen_result}

% \vspace{-5pt}
\noindent\textbf{Effect of retrieved exocentric videos.}
Table~\ref{tab:caption} shows the comparison results on the impact of retrieved samples. We vary the number of exocentric videos $K$ from 0 to 8, 
which is referred to as $K$-shot evaluation~\cite{flamingo}. 
In the 0-shot scenario, where no exocentric videos are used, 
both LLaMA- and MPT-based captioning models achieve relatively low captioning performance, with the MPT-based model achieving only a 0.468 CIDER score. 
In contrast, by leveraging only one exocentric video as reference, 
the model demonstrates a significant improvement across all metrics. 
The captioning performance gradually improves as the number of retrieved samples increases, indicating that the multimodal captioning model successfully leverages relevant visual and textual information from exocentric videos. 
This observation aligns with prior studies on few-shot learning~\cite{gpt3,liu2021makes,yang2022empirical,flamingo}.
Compared with refined caption, using the original ASR transcript as the text input prompt leads to a slight decrease in performance for LLaMA-based captioning model~(CIDER 1.665 vs.~1.768).
We also re-implement a prior egocentric video captioning model on our training set, {\em i.e.,}~LaViLA-GPT2~\cite{lavila}. Our retrieval-augmented captioning model outperforms LaViLA by a large margin.

\begin{table}[t]
\centering
\resizebox{\columnwidth}{!}{
  \begin{tabular}[t]{lccccc}
    \toprule
    Model & Shot & BELU-4 & METEOR & ROUGE-L & CIDER \\
    \midrule
    LaViLA-GPT2~\cite{lavila} & 0 & 0.131&0.287&0.478&0.751 \\
    LaViLA-GPT2-XL & 0 & 0.170 & 0.305 & 0.514 & 1.030 \\
    LaViLA-LLaMA7B & 0 & 0.174 & 0.313 & 0.516 & 1.089\\
    \midrule
    \midrule
    \emph{Ours} \\
    LLaMA-7B & 0 & 0.125&0.276&0.458&0.541\\
    LLaMA-7B-ASR & 1 & 0.277&0.384&0.583&1.665\\
    LLaMA-7B & 1 & \textbf{0.284}&\textbf{0.387}&\textbf{0.589}&\textbf{1.738}\\
    \midrule
    MPT-1B & 0 & 0.120&0.279&0.447&0.468 \\
    MPT-1B & 1 & 0.322&0.400&0.612&1.973 \\
    MPT-1B & 2 & 0.325&0.402&0.618&2.054 \\
    MPT-1B & 4 & \textbf{0.326}&0.404&0.620&2.070\\
    MPT-1B & 8 & 0.320 & \textbf{0.407} & \textbf{0.622} & \textbf{2.107} \\

    \bottomrule
  \end{tabular}
}
\vspace{-5pt}
\caption{Video captioning results on Ego4d validation set (cooking subset). Shot refers to the number of exocentric videos.}
\vspace{-10pt}
\label{tab:caption}
\end{table}

\vspace{4pt}
\noindent\textbf{Performance on EgoLearner.} 
We evaluate the generalisation ability of EgoInstructor by directly transferring the model trained on Ego4d cooking subset to EgoLearner cooking videos. 
As seen from Table~\ref{tab:cap_egolearn}, in the 0-shot scenario, 
our captioning models fail to achieve satisfactory results, 
which can be attributed to the discrepancy between the video and caption distribution of two datasets. 
In the 1-shot scenario, a significant boost can be observed, 
demonstrating that EgoInstructor effectively transfers knowledge from retrieved exocentric videos to assist ego-video captioning, achieving fast adaptation from Ego4d to EgoLearner. 
In addition, we study two variants of retrieved samples: (1) random selection of exocentric videos and (2) ground-truth paired exocentric videos.
Table~\ref{tab:cap_egolearn} reveals that using randomly selected exocentric videos yields similar performance to 0-shot captioning, indicating that random samples do not provide useful semantics for EgoInstructor. 
In contrast, samples retrieved by our cross-view retriever lead to a large gain of captioning performance, and using GT paired exo-videos performs even better. 

\subsubsection{Qualitative Results}
As shown in Fig.~\ref{fig:vis}, we showcase the retrieved third-person videos relevant to the given egocentric videos.
As can be seen from examples, the retrieved samples are generally associated with actions or objects present in the egocentric videos. 
With the assistance of these third-person videos and corresponding captions, the captioning model accurately describes fine-grained objects in the video, such as ``pepper" or ``bowl". 
In contrast, the captioning model without third-person video references generates less accurate descriptions. 
These visualisation results demonstrate the effectiveness of our EgoInstructor in leveraging third-person videos to enhance egocentric video understanding. We have also experimented with the GPT4-Vision model~\cite{gpt4}. Though GPT4-V excels at detecting objects in the video, the generated caption is not precise due to the incorrect perception of interacting objects in egocentric videos.

\begin{table}[t]
\centering
\resizebox{\columnwidth}{!}{
  \begin{tabular}[t]{lccccc}
    \toprule
    Model & Shot & BELU-4 & METEOR & ROUGE-L & CIDER \\
    \midrule
    LaViLA-GPT2~\cite{lavila} & 0 & 0.016	& 0.069 & 0.175 & 0.085 \\
    LaViLA-GPT2-XL & 0 & 0.017	& 0.069	& 0.175	& 0.078 \\
    LaViLA-LLaMA7B & 0 & 0.015 & 0.082 & 0.173 & 0.085 \\
    \midrule
    \midrule
    \emph{Ours} \\
    LLaMA-7B & 0 & 0.005 & 0.071 & 0.140 & 0.082 \\
    LLaMA-7B & 1 & 0.024 & 0.120 & 0.226 & 0.204  \\
    LLaMA-7B (random) & 1 & 0.008 & 0.071 & 0.173 & 0.052 \\
    LLaMA-7B (GT) & 1 &  \textbf{0.029}&\textbf{0.134}&\textbf{0.240}&\textbf{0.251}\\
    \midrule
    MPT-1B & 0 & 0.006 & 0.070 & 0.156 & 0.065\\
    MPT-1B & 1 & 0.024 & 0.130 & 0.241 & 0.245 \\
    MPT-1B (random) & 1 & 0.010 & 0.071 & 0.182 & 0.054 \\
    MPT-1B (GT) & 1 & \textbf{0.032} &\textbf{ 0.153} & \textbf{0.261} & \textbf{0.327}\\
    \bottomrule
  \end{tabular}
}
\vspace{-5pt}
\caption{Zero-shot and one-shot video captioning results on EgoLearner. All the models are trained on Ego4d cooking subset and directly transferred to EgoLearner.}
%\vspace{-0.4cm} 
\vspace{-10pt} 
\label{tab:cap_egolearn}
\end{table}

% \vspace{-5pt}
\section{Conclusion}
In this paper, we consider the problem of egocentric video captioning,
and present a retrieval-augmented model, named EgoInstructor. 
It aims to retrieve semantically similar third-person instructional videos as references and generate captions for egocentric videos. 
We design a cross-view visual representation alignment for retrieval, 
where we train a cross-view retrieval module on automatically curated ego- and exo-video pairs. We propose a novel EgoExoNCE loss for training retrieval module, which aligns ego- and exo-video features to shared text features representing similar action semantics. Across seven benchmarks, our retrieval module demonstrates strong zero-shot cross-view and cross-modal retrieval ability.  With the capability of cross-view retrieval, quantitative and qualitative results show the benefits of leveraging exocentric videos to enhance egocentric video captioning.

\appendix
{\small
\bibliographystyle{ieeenat_fullname}
\bibliography{main}
}

% WARNING: do not forget to delete the supplementary pages from your submission 
% \input{sec/X_suppl}

\end{document}